\documentclass{article}

    \usepackage[final,nonatbib]{neurips_2020}

\usepackage[utf8]{inputenc} 
\usepackage[T1]{fontenc}    
\usepackage{hyperref}       
\usepackage{url}            
\usepackage{booktabs}       
\usepackage{amsfonts}       
\usepackage{nicefrac}       
\usepackage{microtype}      
\usepackage{epsfig}
\usepackage[square,numbers]{natbib}
\usepackage{multirow}
\usepackage{bm}

\title{Building Damage Mapping with Self-Positive Unlabeled Learning}

\author{%
  Junshi Xia \\
  RIKEN\\
  \texttt{junshi.xia@riken.jp} \\
  \And
  Naoto Yokoya \\
  The University of Tokyo and RIKEN \\
  \texttt{yokoya@k.u-tokyo.ac.jp} \\
  \AND
   Bruno Adriano \\
   RIKEN \\
   \texttt{bruno.adriano@riken.jp} \\
}

\begin{document}

\maketitle

\begin{abstract}
Humanitarian organizations must have fast and reliable data to respond to disasters. Deep learning approaches are difficult to implement in real-world disasters because it might be challenging to collect ground truth data of the damage situation (training data) soon after the event. The implementation of recent self-paced positive-unlabeled learning (PU) is demonstrated in this work by successfully applying to building damage assessment with very limited labeled data and a large amount of unlabeled data. The self-PU learning is compared with the supervised baselines and traditional PU learning using different datasets collected from the 2011 Tohoku earthquake, the 2018 Palu tsunami, and the 2018 Hurricane Michael. By utilizing only a portion of labeled damaged samples, we show how models trained with self-PU techniques may achieve comparable performance as supervised learning. 

\end{abstract}

\section{Introduction}
Natural disasters, such as earthquakes, tsunami, and wildfires, are extremely rare occurrences that can wreak havoc on urban environments, resulting in enormous human and economic losses~\cite{unreport}. When a new natural disaster strikes, accurate and rapid actions are necessary in Humanitarian Assistance and Disaster Response (HADR) in order to save thousands of lives and prevent further damage~\cite{DBLP:conf/eccv/WeberMPBLO0020}. Thus, a high-precision disaster damage mapping product needs to meet this demand. However, using very high-resolution optical and SAR images, the disaster damage areas are manually annotated by specialists, which might take a long time if the disaster area is huge~\cite{rs12172839}.

Deep learning is an effective method to automatically recognize damages in remote sensing datasets~~\cite{8126255, Ma2019}. However, deep learning models necessitate a huge number of training images as well as a large amount of high-quality labeled data~~\cite{Everingham2015}. This is not possible in the case of new disasters. One possible solution is to use the datasets and labels of past disasters to identify the damages of new disasters. However, due to domain gaps caused by many factors, such as various geographical locations, different damage characteristics, and seasonal variation, models that have only been trained on previous disasters would not work well for future disasters.

Since human-expert labeling is time-consuming, only a small number of labels are used when annotations obtained after a disaster are used to train models. The availability of labeled data is extremely restricted right after a new disaster, yet an enormous amount of unlabeled satellite images can be acquired over the affected area. \citet{lee2020assessing} applied two semi-supervised learning (SSL) techniques, MixMatch and FixMatch, for the building damages with very limited labeled samples (damaged and undamaged). However, it is very time-consuming to collect both damaged and undamaged samples. Positive-unlabeled learning (PU), which integrates only positive and unlabeled training data, has recently made significant progress~\cite{DBLP:journals/ml/BekkerD20}. By using PU techniques, we can train accurate damage assessment models for new disasters by spending less time on gathering fewer positive labels of damaged class.

In this work, the recent Self-PU learning~\cite{pmlr-v119-chen20b} is applied to train building
damage detection models using very limited labeled positive damage samples. The method is applied to three different datasets to validate the performance. We demonstrate that by utilizing only a fraction of the labeled positive damage, models may obtain outcomes that are comparable to those obtained from fully supervised methods.

\section{Related works}
\textbf{Deep learning in building damage mapping}
Previous studies conducted in fully supervised settings have effectively utilized deep learning algorithms for building damage assessment. The xView2 Challenge~\cite{gupta2019xbd} developed the first large-scale pre- and post-disaster datasets, namely xBD, for mapping building damage, which incorporates VHR optical imagery taken during 19 disasters, including floods, wildfires, and earthquakes. The winning solution adopted a Siamese neural network which is initialized by the building detection from pre-event dataset. \citet{hao2020attentionbased} combined the features from pre- and post-disaster imagery with a Siamese attention-based Unet model. \citet{gupta2020rescuenet} proposed an end-to-end RescueNet for both building segmentation and individual building damage level assessment.
\citet{shen2020crossdirectional} developed a cross-directional fusion strategy to investigate the relationships between the datasets taken before and after the disaster. \citet{boin2020multiclass} proposed the techniques to mitigate the imbalanced class problem for different damage levels. \citet{ADRIANO2021132} construct high-resolution optical imagery and high-to-moderate-resolution SAR data. The multi- and cross-modality deep neural networks are applied to the new dataset. 

In the above studies, the testing datasets are drawn from the same distribution as the training dataset. They do not address the difficulty of conducting inference for a disaster that just happened with very limited labeled datasets. \citet{xu2019building} and~\citet{benson2020assessing} investigated the generalization experiments among the disasters in different regions and indicated that a significant generalization gap existed in current state-of-the-art models. The findings of models trained on previous disasters did not perform well when applied to a new disaster that did not have any labeled datasets.

\textbf{Positive-unlabeled (PU) learning} PU learning is a subdivision of SSL. However, when compared to SSL, there are no negative samples in PU learning, making the task more difficult and necessitating the invention of an algorithm to compensate for the lack of negative samples~\cite{DBLP:journals/ml/BekkerD20}. There are currently two solutions: 1) find credible negative samples in unlabeled data using heuristics or semi-supervised learning approaches~\cite{conf/ijcai/LiL03}, 2) treat unlabeled samples as weighted negative ones, including non-negative PU (nnPU)~\cite{conf/icml/PlessisNS15}. To further improve the performance of PU learning, the self-paced learning~\cite{pmlr-v119-chen20b}, active learning is considered in the framework of PU learning. In the remote sensing applications, \citet{9112285} investigated eigenvalue statistical components-based PU-Learning to extract built-up from PolSAR datasets. \citet{9090858} produced a landslide susceptibility map via the PU-bagging technique.

\section{Data}
\textbf{ABCD.} ABCD (AIST Building Change Detection) dataset~\footnote{https://github.com/gistairc/ABCDdataset}~\cite{7986759} is a new labeled dataset to identify whether buildings have been washed-away by tsunami of Tohoku region of Japan in 2011.
The class label assigned to each patch pair (i.e. "damaged" or "survived") represents whether or not a building at the center of the before-tsunami patch got washed-away by tsunami. In this work, the number of images (the size is $160 \times 160$) for "damaged" and "survived"  are 4253/4253. 

\textbf{xBD.} In additional to ABCD dataset for 2011 Tohoku earthquake, we generated the patch-based datasets for the 2018 \textbf{Palu tsunami} and 2018 \textbf{Hurricane Michael} from xBD. There are four damage levels, including \textit{no damage}, \textit{minor damage}, \textit{major damage} and \textit{destroyed}.  To formulate the binary classification problem and make the class label same as ABCD, 
{major damage} and {destroyed} are categorized into the class \textit{Damaged}, while {no damage},  and {minor damage} are combined as the class \textit{Survived}. The patches of each building were cropped from the original WorldView images of xBD with a fixed size of 128 $\times$ 128 pixels. Thus, the total number of samples for the 2018 \textbf{Palu tsunami} and 2018 \textbf{Hurricane Michael} are 7645 (27\% as the damaged) and 9214 (29\% as the damaged), respectively. 

\section{Proposed approach}
\begin{figure}[!h]
\centering
\includegraphics[width=\textwidth]{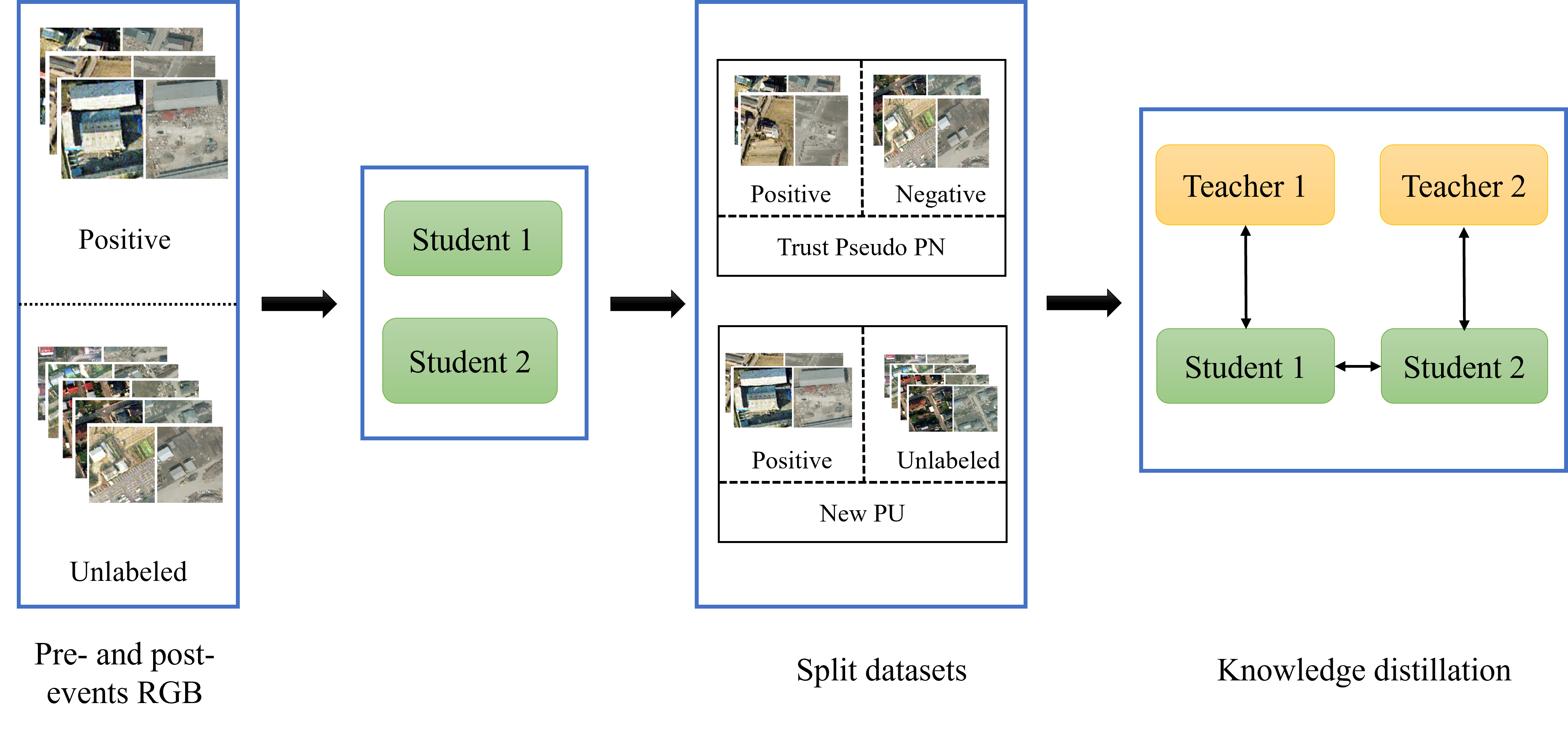}
\caption{The proposed approach.}
\label{fig:proposed_method}
\end{figure}
The objective is formulated as a binary classification with two classes ($\boldsymbol{X}\in \mathcal{R}^{6}$, $Y \in \{-1,+1\}$): damaged and survived, where $\boldsymbol{X}$ is the input that contains six bands of pre- and post-event RGB and $Y$ is the class label. The damaged and survived classes are treated as the positive ($+1$) and negative ($-1$). In the disaster damage mapping, we only have few numbers of damaged samples (P). The remaining datasets are unlabeled (U), which include positive or negative.
Assume that we have the training dataset $D$, which include positive ($D_{P}$) and unlabeled datasets ($D_U$). Then, we have $D=D_P+D_U$, where $D_P$ and $D_U$ contain $n_P$ and $n_U$ samples, respectively. Let $p(x,y)$ be the underlying joint distribution of $(\boldsymbol{X},Y)$, $p(x|Y=+1)$ and $p(x|Y=-1)$ are the distributions of P and N datasets. $\pi_{p}=p(Y=+1)$ and $\pi_{n}=p(Y=-1)=1-\pi_{p}$ are the positive and negative class-prior probabilities. As shown in Fig.~\ref{fig:proposed_method}, we firstly train a classification model to create a trusted pseudo PN dataset for U ($D_{trust}$), and the remaining is treated as a new PU dataset ($D_{U}-D_{trust}$). Then, knowledge distillation with 
two student models and their teacher models is used as a
regularization to enhance performance while overcoming the inconsistent between the two models. 

Due to the fact that deep neural networks are prone to memorizing simple patterns first, followed by irregular patterns, We might start by looking at simple instances and labeling them with confidence. Given an input $x$ with the label $y$ to the given model $g$, the probability of $x$ being positive are calculated. A larger probability implies greater confidence that $x$ belongs to the positive class, and vice versa. Based on this assumption, we selected $n$ most confident positive and negative samples from $D_{U}$ to form $D_{trust}$. Then, the loss function that integrates the supervised one of $D_{trust}$ and PU loss of $D_{U}-D_{trust}$, is given by
\begin{equation}
    L_{hybrid} = \sum_{x \in D_{trust}} L_{CE}(x,y)+\sum_{x \in D_{U}-D_{trust}}L_{nnPU}(x)
\end{equation}
where, the cross-entropy loss are used for $D_{trust}$ and non-negative PU (nnPU)~\cite{conf/icml/PlessisNS15} are used for $D_{U}-D_{trust}$.
To make the the selected confident samples as accurate as possible, We start by training 10 epochs to “warm up” the model. Two ways are used to select the trusted pseudo PN datasets: 1) fixed size: the model choose the fixed number of samples while removing lower-confidence samples, 2) without replacement: the model choose a fixed number of samples without removing lower-confidence samples. In this work, we adapt the similar dynamic selection strategy~\cite{pmlr-v119-chen20b}: linearly increase the size of $D_{trust}$ from 0 to 20\% of $D_u$ with the 75\% bootstrap size of training set for the model.  

Then, knowledge distillation with two student models ($g_1$ and $g_2$) and two teacher models ($G_1$ and $G_2$) are further applied~\cite{pmlr-v119-chen20b}. The two student networks shared the same $D = D_{P}+D_{U}$ and same networks architectures. However, two different trusted pseudo PN datasets may be extracted by two student networks, thus, the consistently between two student networks should be considered as the distillation via MSE loss.  The distillations from $g_1$ to $g_2$ and from $g_2$ to $g_1$ are given by:
\begin{equation}
    L_{MSE}(g_1,g_2,x) = \|f(g_1(x))-f(g_2(x))\|^{2}, x \in D-D_{trust1}
\end{equation}
\begin{equation}
    L_{MSE}(g_2,g_1,x) = \|f(g_2(x))-f(g_1(x))\|^{2}, x \in D-D_{trust2}
\end{equation}
where, $f$ is a monotonic function of mapping (e.g., sigmoid function).
The loss function of the student networks is given by (we only consider the MSE over hard unlabeled samples with larger nnPU):
\begin{equation}
    L_{student}= \sum_{x \in D_{U}-D_{trust1}}L_{S}(g_1,g_2,x)+ \sum_{x \in D_{U}-D_{trust2}}L_{S}(g_2,g_1,x)
\end{equation}
where, $L_{S}(g_1,g_2,x) = L_{MSE}(g_1,g_2,x)$ when $L_{nnPU}(x)>\alpha L_{MSE}(g_1,g_2,x)$, otherwise $L_{S}(g_1,g_2,x) = 0$, $\alpha$ is the control parameter. 

Beside the distillation between two student networks, two teacher networks are included to make sure the consistency with the weighted moving average trajectory of the student networks. The weights of $G_1$ and $G_2$ are updated as
\begin{equation}
   \Theta_{1,t} = \theta_{1,t-1}+\beta\theta_{1,t} 
\end{equation}
\begin{equation}
   \Theta_{2,t} = \theta_{2,t-1}+\beta\theta_{2,t} 
\end{equation}
where, $\Theta_{1,t}$ and $\theta_{1,t}$ are the parameters of $G_1$ and $g_1$ at time $t$, respectively. $\beta$ is the user-defined parameter. The distillations from $G_1$ to $g_1$ and from $G_2$ to $g_2$ are given by:
\begin{equation}
    L_{teacher}= \sum_{x \in D_{U}}\|f(G_1(x))-f(g_1(x))\|^{2}+\sum_{x \in D_{U}}\|f(G_2(x))-f(g_2(x))\|^{2}
\end{equation}
The total loss function is given by:
\begin{equation}
    L = L_{hybrid}+L_{student}+L_{teacher}
\end{equation}


\section{Experimental results and analysis}
\textbf{Settings.} The efficacy of the proposed methods on classifiy damages are tested on ABCD and xView2 datasets. We split the datasets with the ratios of 4/3/3 for train/validation/test. To formulate the PU setting, we select r\% (r ranges from 10 to 30) damaged samples as P, and the remaining (1-r)\% damaged and 100\% survived class are treated as U. The fully-supervised learning with the following setting: 1) Full PN; 2) Standard PN (r\% P+100\% N); 3) Small PN (r\% P+r\% N); 4) PU (U is treated as the N) and nnPU are used as the baselines. In this work, we adopt the simple VGG11 that contains eight convolutional layers and three full-connected layer. The pre- and post-disaster RGB images are treated as the input of the models.
We set the SGD with momentum of 0.9 and a batch size of 16 for 100 epochs on a single NVIDIA V100 GPU with 16GB memory.

\begin{table}[!h]
\caption{Classification accuracy of the proposed method and the baselines averaged over 5 runs}
\label{table:accuracies}
\centering
\begin{tabular}{p{1.5cm}p{1.5cm}cccccc}
\toprule
\textbf{Datasets}     & \textbf{Ratio (\%)}&\textbf{Full PN} & \textbf{Standard PN} & \textbf{Small PN} & \textbf{PU} & \textbf{nnPU} & \textbf{Self-PU}\\
\midrule
\multirow{3}{*}{ABCD} & 10&\multirow{3}{*}{0.94} & 0.72 & 0.64 & 0.53 & 0.74 & 0.79  \\
        & 20& & 0.75 & 0.67 & 0.54 & 0.78 & 0.81  \\
        & 30& & 0.77 & 0.69 & 0.55 & 0.80 & \textbf{0.87}  \\
\midrule
\multirow{3}{*}{Palu tsunami} & 10&\multirow{3}{*}{0.95} & 0.74 & 0.71 & 0.64 & 0.79 & 0.82  \\
         & 20&& 0.76 & 0.73 & 0.66 & 0.82 & 0.87  \\
         & 30&& 0.81 & 0.75 & 0.66 & 0.85 & \textbf{0.89}  \\
\midrule
Hurricane & 10&\multirow{3}{*}{0.92} & 0.76 & 0.67 & 0.59 & 0.79 & 0.82  \\
  Michael   & 20&& 0.78 & 0.69 & 0.60 & 0.84 & 0.84  \\
         & 30&& 0.80 & 0.72 & 0.59 & 0.86 & \textbf{0.86}  \\
    \bottomrule
  \end{tabular}
\end{table}
\textbf{Results.} Table~\ref{table:accuracies} reports the accuracies of different methods. The fully-supervised model on all labeled data (Full PN), which is not available in real scenario, can be treated as the upper bound of the performance. The fully-supervised model with PU setting obtained the worst results because the training label is very noisy, especially for the negative labels. The clean training sets with small amount of P (standard PN and small PN) improved the results. nnPU makes further improvements in all cases. Self-PU improves the performance on all three disaster datasets. It should be noticed that we can obtain nearly the accuracies of 90\% using only 30\% positive damaged class as the input. 

\begin{table}[!h]
  \caption{Classification accuracies with different components on ABCD datasets.}
  \label{table:accuracies_coponment}
  \centering
 \large
  \begin{tabular}{cccc}
    \toprule
    \textbf{Methods}     & \textbf{Accuracy} & \textbf{Methods} & \textbf{Accuracy} \\
    \midrule
        nnPU & 0.74 &           $L_{hybird}$ & 0.76 \\ 
         $L_{hybird}$ (fixed size) & 0.75 & $L_{hybird}+L_{student}$ & 0.77\\
         $L_{hybird}$ (w.o. replacement) & 0.74 & Self-PU & \textbf{0.79}   \\
 
    \bottomrule
  \end{tabular}
\end{table}
\begin{table}[!h]
  \caption{Sensitive analysis of $\alpha$ and $\beta$ on the performance of ABCD dataset.}
  \label{table:accuracies_parameters}
  \centering
 \large
  \begin{tabular}{cccc}
    \toprule
    $\alpha$     & \textbf{Accuracy} & $\beta$ & \textbf{Accuracy} \\
    \midrule
         10 & 0.76 & 0.3 & 0.77 \\ 
         \textbf{20}  & \textbf{0.79} & 0.4 & 0.77\\
         30  & 0.77 & \textbf{0.5} & \textbf{0.79}   \\
         40  & 0.75 & 0.5 & 0.78   \\ 
    \bottomrule
  \end{tabular}
\end{table}

\textbf{Ablation studies.} From Table~\ref{table:accuracies_coponment}, it can be clearly seen that the proposed Self-PU methods gain the better performance than the ones which only consider the $L_{hybird}$ and $L_{hybird}+L_{student}$. Table~\ref{table:accuracies_parameters} investigates three different values of $\alpha$ and $\beta$. $\alpha$ indicates the quality of consistency between two students and $\beta$ controls the updates of teachers from the students.
Both parameters favors a reasonably moderate values. Larger or smaller values will decrease the accuracies.

\section{Conclusion}
In this paper, we introduced a novel application of PU learning to detect
damaged buildings in satellite imagery with limited labeled samples. We experimented with the recent
PU techniques with the combination of self-paced learning and knowledge distillation, and showed how they are able to achieve strong performance using very limited samples of damaged buildings by leveraging unlabeled samples. They consistently outperformed fully
supervised models and even achieved performance close to that of a fully supervised setting with no
data constraints. The results empirically showed how self-PU learning approaches can be useful to train models
when a new disaster is unfolding in an unseen region. For future work, we will work towards the combination of the labeled datasets from past disasters and very limited damaged samples to further improve the performance.  
.

\section{Application Context}
In the HADR domain, the most urgent need is to provide the building damage mapping that requires a large number of labeled datasets as the input of current SOTA deep learning models, which are already explored in the past AI for HADR workshops~\cite{shen2020crossdirectional, boin2020multiclass}. However, such labeled datasets are not available for a new disaster. Previous studies in~\cite{xu2019building} and~\cite{benson2020assessing} proved that the result of a new disaster is not accurate by training the model on the past disasters. \citet{lee2020assessing} attempted to apply SSL methods for the building damage mapping. In our work, we provided the building damage mapping using very limited positive damaged images, requiring less time for the annotation before the response. 

\pagebreak
\section*{Acknowledgments and Disclosure of Funding}
This work was supported by KAKENHI (Grant Number 19K20309 and 19H02408), JSPS Bilateral Joint Research Project (Grant number JPJSBP 120203211).
\small
\bibliographystyle{abbrvnat}
\bibliography{refs}

\end{document}